\documentclass[sigconf, nonacm]{acmart}

\setcopyright{none}
\acmISBN{}
\acmDOI{}
\title[Centerline-Guided Road Surface Segmentation]{Centerline-Guided Multi-Class Road Surface Segmentation}

\author{Xu Ouyang}
\affiliation{%
  \institution{HERE North America}
  \country{United States}
}
\author{Lin Gan}
\affiliation{%
  \institution{HERE North America}
  \country{United States}
}
\author{Di Ma}
\affiliation{%
  \institution{HERE North America}
  \country{United States}
}
\author{Sonnie Chen}
\affiliation{%
  \institution{HERE North America}
  \country{United States}
}
\author{Sanjay Boddhu}
\affiliation{%
  \institution{HERE North America}
  \country{United States}
}

\begin{abstract}
Road-surface extraction from satellite imagery supports geographic
mapping, but a binary mask leaves road axes, transverse extent, and
junction structure implicit. Learning these properties together is
challenging because narrow corridors, merging roads, and occlusions
require different geometric cues. We present a centerline-guided
multi-class segmentation framework that makes these dependencies
explicit. A hierarchical encoder supplies shared features, while
predicted centerline features and soft probabilities condition
specialized regular, intersection, and overpass decoders.
A two-component width branch supplies transverse geometry to
regular-surface decoding, and learned feature fusion produces the
complete road mask. Unlike independent prediction heads or a fixed
union of component masks, this design exchanges geometric information
during decoding while retaining separately inspectable outputs.
Evaluation of the four-semantic-output configuration on CHN6-CUG
achieves 65.53\% complete-surface IoU, improving on a binary
SegFormer-B4 baseline by 1.01 percentage points and reducing
false-positive pixels by 11.94\%. The results support geometry-aware
surface refinement and a richer raster representation for road-map
inspection.
\end{abstract}

\ccsdesc[500]{Computing methodologies~Image segmentation}
\ccsdesc[300]{Information systems~Geographic information systems}

\keywords{road extraction, satellite imagery, multi-task learning,
centerline guidance, width estimation, geographic information systems}

\begin{document}
\maketitle
\raggedbottom

\section{Introduction}

Road maps support infrastructure inventory, accessibility analysis,
and network updating. These applications use related representations:
surface masks describe extent, centerlines summarize longitudinal
structure, and width describes transverse geometry. At intersections,
corridors merge and cannot be described as a single narrow strip.
Joint outputs expose this geometry rather than only a binary raster.

\begin{figure}[t]
  \centering
  \includegraphics[width=\columnwidth]{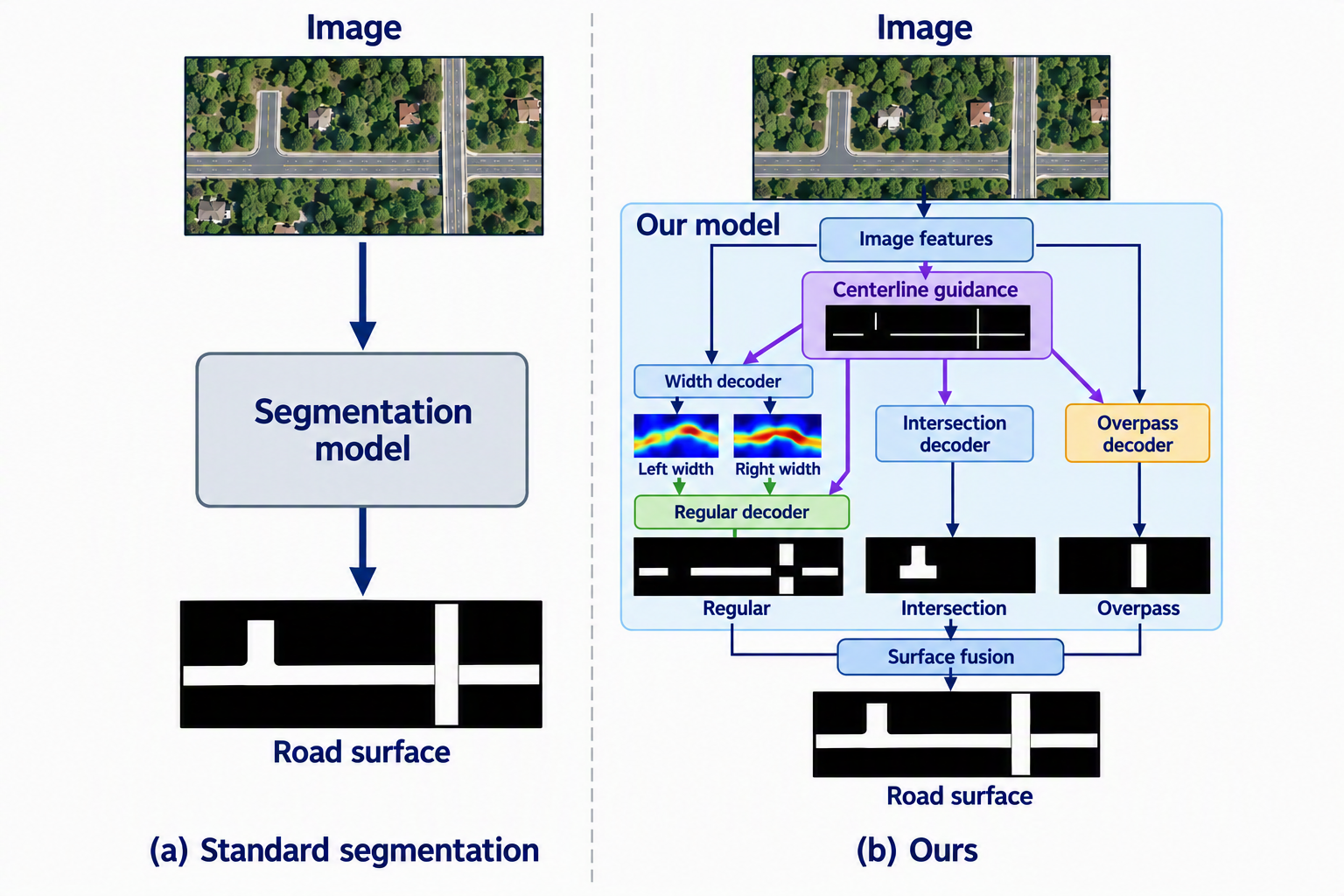}
  \caption{From binary road extent to structured road representations.
  Conceptual comparison of standard surface segmentation with our
  centerline-guided regular, intersection, and overpass branches,
  two-sided width estimation, and surface fusion.}
  \Description{Conceptual comparison of standard road-surface
  segmentation with centerline-guided multi-branch prediction,
  two-sided widths, and final surface fusion.}
  \label{fig:teaser}
\end{figure}

Narrow roads can be confused with their surroundings or interrupted
by vegetation and shadows. Centerlines are sensitive to displacement,
while width requires a plausible location and directional consistency.
This motivates sharing geometric predictions during surface decoding.

A SegFormer-B4 encoder~\cite{xie2021segformer} supplies shared features.
Predicted centerline features and probabilities condition regular,
intersection, and overpass branches. Width features and
local left- and right-side widths further guide regular surface. Learned
refinement combines branch features and logits into a complete mask,
rather than a fixed union. Figure~\ref{fig:teaser} contrasts binary
extent with our structured prediction framework.

The contributions of this study are:
\begin{itemize}
  \item \textbf{Centerline-conditioned surface decoding.}
  Predicted centerline features and soft probabilities guide downstream
  surface branches, making longitudinal road structure available during
  decoding rather than only through an auxiliary loss or post-processing.
  \item \textbf{Two-sided width-guided regular-surface prediction.}
  Width estimates and latent width features jointly condition the regular
  branch, coupling transverse extent with centerline guidance instead of
  treating width as an independent output.
  \item \textbf{Complementary-region modeling with learned refinement.}
  Specialized regular, intersection, and overpass branches retain
  region-specific representations. Feature-and-logit fusion combines
  them into complete surface without imposing mutually exclusive
  classes or a fixed mask union.
\end{itemize}

On the public CHN6-CUG dataset~\cite{zhu2021global,chn6official},
paired evaluation against identical complete-surface targets shows an
IoU increase from 64.53\% to 65.53\%, supported by class-wise measurements
and qualitative analysis.

\section{Related Work}

\subsection{Dense Road Surface Extraction}

Encoder--decoder models combine contextual recognition with spatial
localization. U-Net introduced a contracting path and an expanding path
for precise dense prediction~\cite{ronneberger2015unet}. Although developed
for biomedical images, this design also underlies road-oriented models
such as the stacked U-Nets of Sun et al.~\cite{sun2018stacked}.
For roads, the central difficulty is not simply identifying pavement:
parking areas, roofs, and construction sites can resemble road
surfaces, while shadows and vegetation can obscure genuine corridors.
The output must combine broad context with localized boundaries.

D-LinkNet addresses road extraction with a pretrained encoder and
dilated convolutions that enlarge the receptive field without reducing
feature-map resolution~\cite{zhou2018dlinknet}. DeepLabv3+ combines
multi-scale atrous context with an encoder--decoder refinement
module~\cite{chen2018deeplab}. These approaches illustrate complementary
roles for contextual aggregation and spatial recovery. GCB-Net further
investigates global context and batch-independent road extraction from
very-high-resolution imagery~\cite{zhu2021global}. Our fusion stage
likewise aggregates spatial scales, but the decoder separates the
geometric cues used for subsequent surface prediction.

\subsection{Hierarchical Features and Structured Decoding}

SegFormer combines a hierarchical Transformer encoder with an efficient
multi-scale segmentation decoder~\cite{xie2021segformer}. Hierarchical
features are attractive for roads because local boundary evidence and
larger-scale corridor context occur at different resolutions.
We use its MiT-B4 encoder as a common backbone for the binary baseline
and proposed model. The methodological focus is how the decoder exposes
and reuses road-specific representations rather than a new Transformer
block.

Independent heads share a backbone but need not exchange predictions.
Our decoder instead exposes centerline features and soft probabilities
to later tasks, and width features and estimates to regular surface.
This separates generic multi-scale fusion from geometric conditioning.

\subsection{Multiple Outputs and Geometric Supervision}

Sun et al. investigate stacked U-Nets with multiple outputs, hybrid
supervision, and post-processing for road
extraction~\cite{sun2018stacked}. Batra et al. jointly learn road
orientation and segmentation with a stacked multi-branch module, then
apply connectivity refinement~\cite{batra2019orientation}. These studies
motivate sharing information between road-related tasks and distinguishing
an auxiliary target's meaning from the mechanism that incorporates it.

Unlike orientation supervision, our decoder couples an explicit
centerline with two width components and specialized surface branches.
Their independently supervised masks may overlap, and learned
refinement combines them rather than imposing a fixed union.

\subsection{Road Graphs and Connectivity}

Raster segmentation and road-graph extraction address related but
different mapping requirements. DeepRoadMapper begins with segmentation
and reasons about missing connections using a shortest-path
formulation~\cite{mattyus2017deeproadmapper}. RoadTracer instead constructs
a graph through iterative exploration guided by a learned decision
function~\cite{bastani2018roadtracer}. VecRoad combines point-based
exploration with segmentation and junction cues to improve alignment
and connectivity~\cite{tan2020vecroad}. Sat2Graph encodes a road graph
as a tensor, enabling prediction of graph structure with a
non-recurrent model~\cite{he2020sat2graph}.

More recent methods further connect dense features and graph prediction.
RNGDet++ adds instance-segmentation supervision and multi-scale backbone
features to road-graph detection~\cite{xu2023rngdet}. SAM-Road adapts a
foundation-model encoder to road/intersection masks and predicts edge
existence with a lightweight graph network~\cite{hetang2024samroad}.
Both reinforce the value of structured intermediate representations,
while targeting vector graphs rather than width-aware road surfaces.

Unlike graph prediction, our outputs are dense masks and width maps.
Surface overlap and centerline agreement assess their geometry,
not route quality or graph connectivity.

\subsection{Thin Structures, Imbalance, and Evaluation}

Thin centerlines make exact overlap sensitive to displacement.
The clDice measure couples masks with their morphological skeletons
and provides differentiable supervision~\cite{shit2021cldice}.
We instead supervise an explicit centerline head with cross-entropy
and Dice and report exact and tolerant agreement.

Class imbalance also affects dense road extraction. Focal loss reduces
the relative contribution of easy examples~\cite{lin2017focal}; we use
it in the binary baseline, while the proposed model combines weighted
binary cross-entropy, Dice, and sparse width regression.
Benchmarks such as DeepGlobe establish task-specific data and evaluation
protocols~\cite{demir2018deepglobe}. Here, all comparisons use a common
CHN6-CUG complete-surface target. Published results on other datasets
are methodological context, not directly comparable scores.

\begin{figure*}[t]
  \centering
  \includegraphics[width=0.94\textwidth]{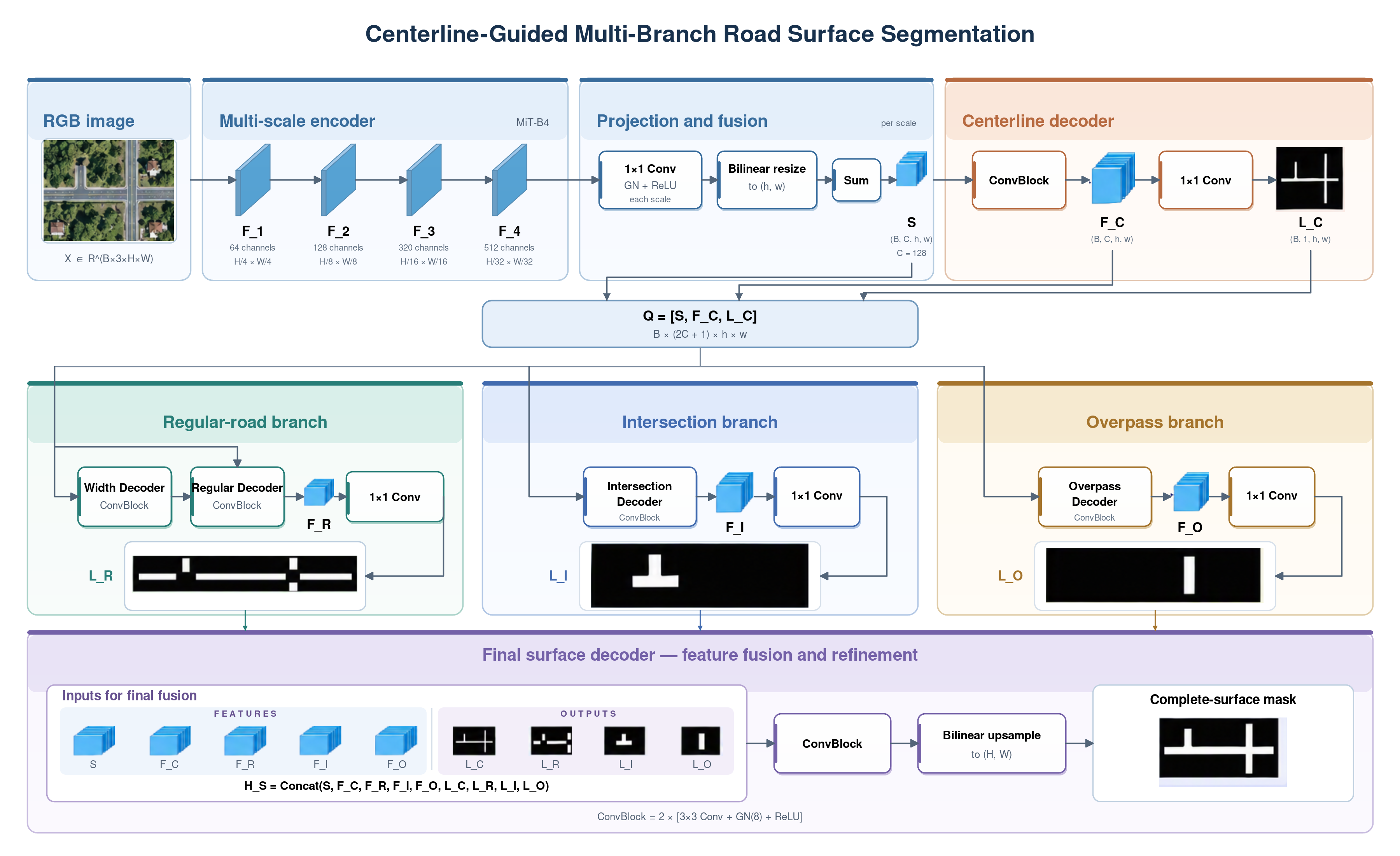}
  \caption{Centerline-guided multi-branch architecture.
  Shared multi-scale features and predicted centerline information form
  $Q$. Width features guide the regular-surface decoder; intersection
  and overpass branches decode complementary regions.
  Final refinement fuses shared features, centerline features and
  probability, and all surface-branch features and logits.}
  \Description{Final supplied architecture with MiT-B4 encoding,
  projected feature fusion, centerline guidance, two-sided width,
  regular, intersection and overpass branches, and final surface fusion.}
  \label{fig:architecture}
\end{figure*}

\begin{figure*}[!t]
  \centering
  \includegraphics[width=0.94\textwidth]{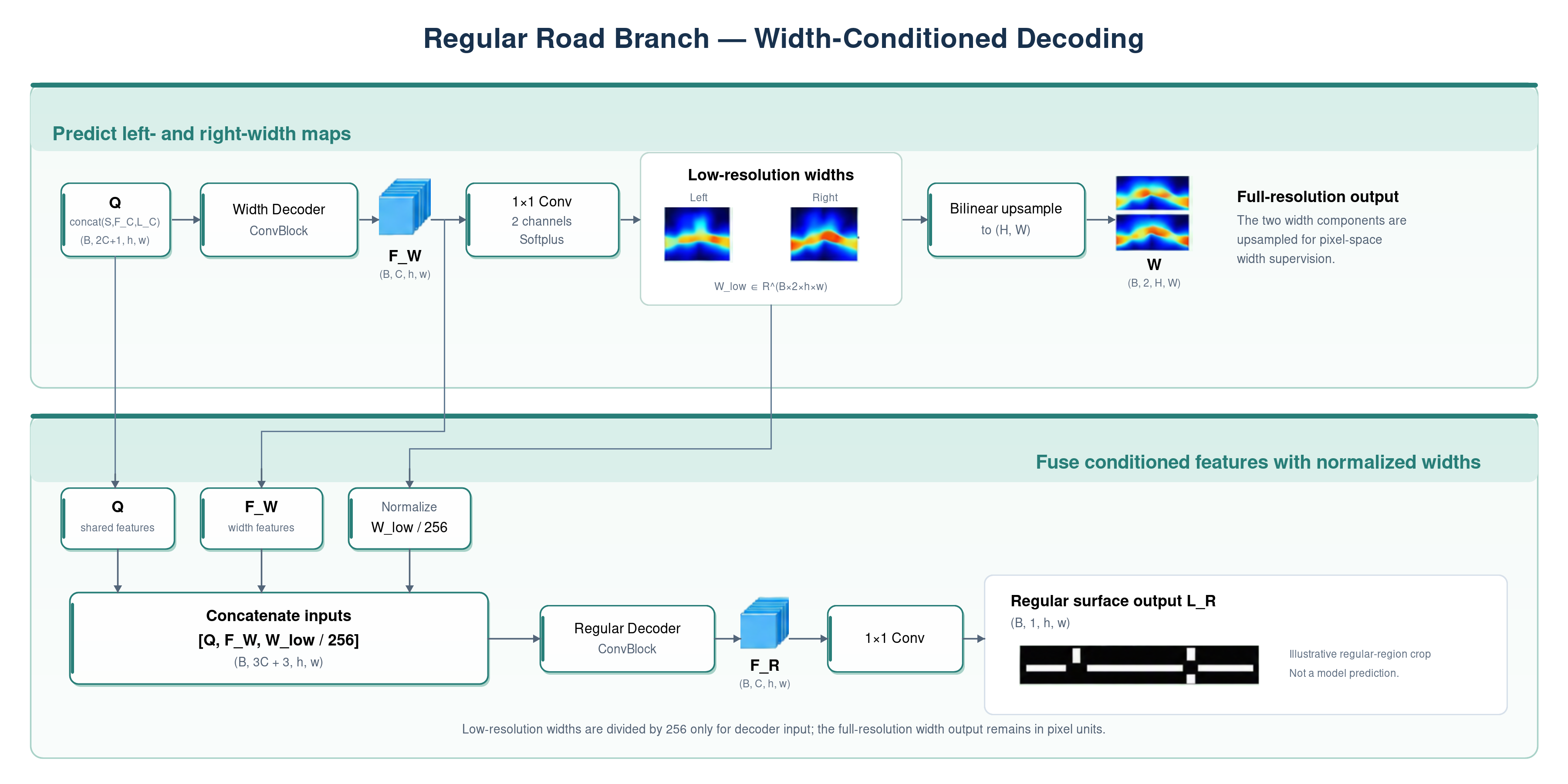}
  \caption{Width-conditioned regular-road branch. The width decoder
  predicts local left- and right-side widths, which are upsampled for
  pixel-space supervision. Low-resolution widths are normalized only
  for concatenation with $Q$ and $F_W$ before regular-surface decoding.
  The crop in the output panel is illustrative, not a model prediction.}
  \Description{Regular-road branch diagram showing local left- and
  right-side width prediction, full-resolution outputs, normalized width
  inputs, and the regular-surface decoder.}
  \label{fig:regular-branch}
\end{figure*}

\section{Centerline-Guided Model}
\label{sec:model}

\subsection{Problem Formulation}

For an RGB image $X\in\mathbb{R}^{H\times W\times3}$, let
$Y_S,Y_C,Y_R,Y_I,Y_O\in\{0,1\}^{H\times W}$ denote complete surface,
centerline, regular surface, intersection surface, and overpass regions.
These are binary
target planes, not the categories of one softmax classifier. In
particular, road extent and centerline describe overlapping aspects of
the same scene. Here, \emph{multi-class} denotes these semantic outputs,
implemented as overlapping binary masks within a multi-task model,
not mutually exclusive softmax categories.

Width supervision adds a validity mask $V$ and two component maps
$W_1,W_2$. The full architectural target representation is
\begin{equation}
  Y=[Y_S,Y_C,Y_R,Y_I,Y_O,V,W_1,W_2].
\end{equation}
The architecture provides five semantic probability maps and two local
side-width maps in one forward pass. Width errors are evaluated only where
$V=1$, not over unannotated background. Complete surface is the primary
comparison target; the remaining outputs expose complementary road
geometry and provide intermediate supervision. Dataset details and
derived-target provenance appear in Section~\ref{sec:experiments}.

\subsection{Hierarchical Encoding and Shared Fusion}

MiT-B4 produces four feature maps $E_\ell$, with channel dimensions
$(64,128,320,512)$ and spatial strides $(4,8,16,32)$.
The configured stage depths are $(3,8,27,3)$. Let $\phi_\ell$ denote a
$1\times1$ projection followed by GroupNorm and ReLU, and let
$\mathcal{U}_4$ bilinearly resize a feature map to stride 4. We form
\begin{equation}
  S=\sum_{\ell=1}^{4}\mathcal{U}_4(\phi_\ell(E_\ell)),
  \label{eq:fusion}
\end{equation}
where each projected feature has 128 channels. A ConvBlock denotes
two $3\times3$ convolutions, each followed by eight-group GroupNorm
and ReLU. The projections align feature dimensions before addition.
The experimental decoder additionally refines the summed features
with a ConvBlock before task decoding.

All decoders use the same 128-channel stride-4 grid. Output resizing
uses bilinear interpolation; \texttt{align\_corners} is false.

\subsection{Centerline-Conditioned Representation}

The centerline decoder first computes a 128-channel feature map
$F_C=\mathcal{B}_C(S)$ and a single-channel soft centerline output
$L_C=\sigma(\psi_C(F_C))$ through a $1\times1$ classifier and sigmoid.
We use $F$ for feature tensors and $L$ for branch outputs. Specifically,
$L_C$ denotes the post-sigmoid centerline probability, while the
surface outputs $L_R$, $L_I$, $L_O$, and $L_S$ denote logits.
The downstream representation is
\begin{equation}
  Q=\operatorname{concat}(S,F_C,L_C).
  \label{eq:condition}
\end{equation}
$Q$ has 257 channels and exposes centerline features and soft confidence,
without thresholding intermediate information.

Training and inference both condition on the predicted centerline,
not ground truth. Neither $F_C$ nor $L_C$ is detached, allowing
downstream losses to propagate through these paths.

\subsection{Local Left- and Right-Side Width Prediction}

The width branch produces features $F_W=\mathcal{B}_W(Q)$ and two
nonnegative outputs:
\begin{equation}
  \widehat{W}^{(4)}
    =\operatorname{Softplus}(\psi_W(F_W)).
\end{equation}
At valid target locations, the two values are distances from a regular-road
centerline to its boundaries along opposite sides of the local normal.
Under the target generator's path-orientation convention, $W_1$ is the
local left-side distance along the positive normal and $W_2$ is the local
right-side distance along the negative normal. Left and right are local
labels tied to the oriented centerline, not compass directions; reversing
the path orientation swaps the side labels, so the two channels retain
their stored order. A centerline with these two side widths describes a
locally varying-width regular-road strip. It does not by itself define
or reconstruct the complete road-surface mask: intersections and overpass
regions are handled by separate surface branches and learned fusion.

Softplus produces positive predictions without a hard clipping boundary.
The stride-4 prediction is bilinearly resized to $H\times W$ for
regression. Its numeric values remain image-pixel widths: interpolation
changes the sampling grid, not the unit of the regressed target. Thus,
there is no additional factor of four applied to the output values.

The same low-resolution width estimates are divided by the configured
256-pixel scale before being provided to the regular-surface decoder.
This normalization controls their input scale; it is not a hard cap on
the Softplus output and does not rescale the width regression target.
The branch predicts the two stored components in their supplied order.
The evaluated objective does not minimize over permutations of the two
channels, so component ordering is part of the target definition.

\subsection{Surface Branches and Final Refinement}

The regular-surface branch combines centerline conditioning with width
features and normalized width maps:
\begin{equation}
  F_R=\mathcal{B}_R\left(
    [Q,F_W,\widehat{W}^{(4)}/256]\right),\quad L_R=\psi_R(F_R).
\end{equation}
Figure~\ref{fig:regular-branch} details this dataflow. Its 387-channel
input combines latent width context with explicit geometric estimates.

The intersection and overpass branches receive $Q$:
\begin{equation}
  F_k=\mathcal{B}_k(Q),\qquad L_k=\psi_k(F_k),\quad k\in\{I,O\}.
\end{equation}
The intersection branch represents merged junction regions. The
overpass branch represents road regions where grade-separated crossings
complicate the surface appearance. Both branches retain
their own features and logits and do not receive the extra width
features or maps.

The complete-surface decoder uses the joint representation
\begin{equation}
  H_S=[Q,F_R,F_I,F_O,L_R,L_I,L_O],\quad L_S=\mathcal{D}_S(H_S).
  \label{eq:final}
\end{equation}
$H_S$ has $5C+4=644$ channels for $C=128$. $\mathcal{D}_S$ consists of a two-convolution
block, Dropout2D with rate 0.1, and a $1\times1$ classifier. Branch logits
are concatenated before thresholding. Consequently, refinement can use
both their confidence and their feature representations, and is not
restricted to a Boolean union or intersection of predicted masks.

The five full-resolution outputs are independently transformed by a
sigmoid. No hard constraint enforces inclusion of the centerline in the
surface, compatibility between the two width components, or exact
agreement between the complete mask and the surface sub-branches.
Those are possible consistency objectives, not features of the
evaluated implementation.

\section{Training Objectives}

\subsection{Weighted Semantic Supervision}

For each semantic task $k$, let $p_i=L_{C,i}$ for the centerline task
and $p_i=\sigma(L_{k,i})$ for surface tasks, and let $y_i$ be
the corresponding binary label. Weighted binary cross-entropy is
\begin{align}
  \mathcal{L}^{\mathrm{BCE}}_k=-\frac{1}{N}\sum_i
  \big[&a_k y_i\log p_i \nonumber\\
       &+(1-y_i)\log(1-p_i)\big],
\end{align}
with positive task weights $a_k$. The evaluated surface and intersection
weights are 8; centerline and regular-surface weights are 4.
These weights affect positive terms, not both sides of the binary
classification problem. The soft Dice term uses smoothing constant 1:
\begin{equation}
  \mathcal{L}^{\mathrm{Dice}}_k
  =1-\frac{2\sum_i p_i y_i+1}
             {\sum_i p_i+\sum_i y_i+1}.
\end{equation}
The implementation computes this overlap per image and averages across
the batch. Each semantic objective is
$\mathcal{L}_k=\mathcal{L}^{\mathrm{BCE}}_k+
\mathcal{L}^{\mathrm{Dice}}_k$.

Cross-entropy supplies local gradients; Dice measures soft overlap.
Outer task coefficients balance the objectives.

\subsection{Sparse Width Regression}

Let $\Omega_V=\{i:V_i=1\}$. Direct regression averages Smooth L1
over both components at valid pixels:
\begin{equation}
  \mathcal{L}_{\mathrm{reg}}
  =\frac{\sum_{i\in\Omega_V}\sum_{d=1}^{2}
      \rho(\widehat{W}_{d,i}-W_{d,i})}
      {\max(2|\Omega_V|,1)},
\end{equation}
where $\rho(e)=e^2/2$ for $|e|<1$ and $|e|-1/2$ otherwise.
The mask prevents missing width annotations from acting as zero-valued
targets. An image without valid width labels contributes zero direct
width error, rather than a fictitious background regression loss.

The additional smoothness term is directional and target-gated. A local
$5\times5$ centerline neighborhood supplies coordinate moments. The
principal-direction angle is
\begin{equation}
  \theta_i=\frac12\operatorname{atan2}
       (2\Sigma_{xy,i},\Sigma_{xx,i}-\Sigma_{yy,i}),
\end{equation}
giving a unit tangent $t_i=(\cos\theta_i,\sin\theta_i)$. For neighboring
centerline pixels $i,j$ that both have valid widths, let $d_{ij}$ be
their unit displacement and let
$b_{ij}=|\langle\operatorname{norm}(t_i+t_j),d_{ij}\rangle|$.
The four tested offsets are $(0,1),(1,0),(1,1),(1,-1)$.

With total annotated width $u_i=W_{1,i}+W_{2,i}$, the pair weight is
\begin{equation}
  \omega_{ij}
    = \mathbf{1}_{\mathrm{valid}(i,j)}
      b_{ij}^{4}\mathbf{1}_{b_{ij}\geq0.8}
      \exp\left(-4\frac{|u_i-u_j|}
        {\max((u_i+u_j)/2,1)}\right).
\end{equation}
Alignment favors variation along a road rather than arbitrary
neighbor directions. The exponential gate reduces the smoothness
penalty near changes already present in the annotated width. The
normalized term is
\begin{equation}
  \mathcal{L}_{\mathrm{smooth}}
  =\frac{\sum_{(i,j)}\omega_{ij}
              \|\widehat{W}_i-\widehat{W}_j\|_1}
         {\max(2\sum_{(i,j)}\omega_{ij},1)}.
\end{equation}
Ground-truth geometry determines these training-loss weights,
not forward-pass inputs at inference.

\subsection{Joint Objective}

The width objective and the architecture's joint loss are
\begin{align}
  \mathcal{L}_W &=
       \mathcal{L}_{\mathrm{reg}}+
       0.05\mathcal{L}_{\mathrm{smooth}},\\
  \mathcal{L} &=\mathcal{L}_S+
       \sum_{k\in\{C,R,I,O\}}\lambda_k\mathcal{L}_k+
       \lambda_W\mathcal{L}_W.
\end{align}
Complete surface has unit weight because it is the main comparison
target. The additional losses supervise interpretable intermediate
products while maintaining complete surface as the primary target.
The evaluated run uses $\lambda_C=\lambda_R=\lambda_I=0.3$ and
$\lambda_W=0.2$ for its supervised outputs.

\section{Experimental Protocol}
\label{sec:experiments}

\subsection{Dataset and Target Preparation}

CHN6-CUG is a public dataset released by the China University of
Geosciences team with GCB-Net~\cite{zhu2021global,chn6official}.
It contains 4,511 Google Earth images of $512\times512$ pixels at
0.5 m/pixel: 3,608 training and 903 test images. Its six urban areas
are Chaoyang in Beijing, Yangpu in Shanghai, central Wuhan, Nanshan
in Shenzhen, Sha Tin in Hong Kong, and Macao. We follow the published
train/test partition. The release restricts the imagery to
non-commercial research and educational use.

Besides the public imagery and complete-surface labels, our experiments
use locally derived centerline, regular-surface, intersection, and sparse
width targets. These auxiliary labels are not part of the public release.
The target-generation procedure remains necessary documentation for a
reproducible training release.

The published train/test identifiers are disjoint. The local test labels
contain 13,563,927 complete-surface pixels out of 236,716,032 pixels (5.73\%).
Centerline, regular, intersection, and valid-width counts are
481,069, 12,368,452, 1,195,475, and 465,040, respectively.
These counts describe the supplied local representation, not a new
published dataset statistic.

\subsection{Training and Initialization}

The available checkpoint evaluates the four-output configuration
$(S,C,R,I)$ with width regression; it uses neither the overpass branch
nor its labels. Its final fusion has 515 channels, and all reported
scores and prediction figures refer to this configuration, not to
an evaluated five-output system.

The multi-task model uses AdamW~\cite{loshchilov2019adamw} with base
learning rate $10^{-5}$, weight decay 0.01, and a tenfold decoder
learning-rate multiplier. Batch size is one with four-step gradient
accumulation. A 1,000-iteration linear warm-up precedes polynomial
decay with power 1 over a configured 160,000-iteration schedule.
The seed is 663097139, and evaluation/checkpoint intervals are 8,000
iterations. Logs record an NVIDIA GeForce RTX 2080 Ti, PyTorch 1.13.1,
and CUDA 11.7.

Training crops are $512\times512$. Augmentation includes random
shadowing, horizontal flips, rotations within $\pm20^\circ$, and
photometric distortion. RGB normalization means are 123.675, 116.28,
and 103.53; standard deviations are 58.395, 57.12, and 57.375.
The preprocessor applies the configured BGR-to-RGB conversion.
The width-aware rotation transform handles multi-channel road targets.
Inference uses whole-image processing rather than test-time ensembling.

The multi-task run initializes through a compatibility hook from its
MGP/APAC checkpoint at iteration 128,000. The selected CHN6-CUG
checkpoint is from iteration 96,000. The binary baseline is a separate
two-class SegFormer-B4 model with a 256-channel decoder, focal
cross-entropy ($\gamma=2$; background/foreground weights 1 and 8), and
foreground Dice weight 0.5. It initializes from its own MGP/APAC
checkpoint at iteration 64,000. The configurations share the optimizer
family, base schedule, crop size, seed, and broad augmentation families,
but not the heads, objectives, initial weights, or target loaders.

\paragraph{Checkpoint selection.}
We use the published train/test partition, but the provided training
configurations also monitor its 903-image test partition for checkpoint
selection (complete-surface F1 for multi-task and foreground IoU for
binary). The reported numbers are therefore test-partition scores
exposed to model selection, not an untouched held-out estimate.

\subsection{Evaluation}

The multi-task operating threshold is fixed at 0.8, selected using
training data. The binary baseline uses 0.5. These settings are fixed
for all reported metrics and visualizations.

For each target, pixel counts are summed over the entire 903-image
partition before computing
\begin{align}
 \mathrm{IoU}&=\frac{TP}{TP+FP+FN},&
 \mathrm{F1}&=\frac{2TP}{2TP+FP+FN},\\
 \mathrm{Precision}&=\frac{TP}{TP+FP},&
 \mathrm{Recall}&=\frac{TP}{TP+FN}.
\end{align}
These micro-aggregated scores are not averages of per-image scores.
All semantic results in the two tables are percentages.

For tolerant centerline evaluation, let $\mathcal{D}_2$ be a
$5\times5$ square dilation, equivalent to a two-pixel tolerance in
the $L_\infty$ grid distance. Tolerant precision is
$|P_C\cap\mathcal{D}_2(Y_C)|/|P_C|$, and tolerant recall is
$|Y_C\cap\mathcal{D}_2(P_C)|/|Y_C|$. Their harmonic mean is reported
separately. This rule allows multiple nearby pixels to match and is
neither one-to-one matching nor a route-connectivity metric.

Width MAE is the absolute error summed over both components and all
valid pixels, divided by twice the valid-pixel count. It is independent
of the semantic threshold. At the source dataset's nominal 0.5 m/pixel,
2.57 pixels correspond to approximately 1.28 m per component, but this
conversion is not a field-survey accuracy estimate or the error of the
sum of the two widths.

\subsection{Paired Prediction Audit}

We aligned saved predictions by image identifier, applied the fixed
thresholds, and verified pixelwise equality of the paired complete-surface
targets. Recomputed pooled metrics reproduce all four reported
complete-surface scores.

For tile $n$, define the paired difference in percentage points:
\begin{equation}
  \Delta_n=100\left(\mathrm{IoU}^{\mathrm{multi}}_n-
                 \mathrm{IoU}^{\mathrm{binary}}_n\right).
\end{equation}
We record per-image confusion counts, IoU, F1, precision, and recall in
an accompanying CSV. For an empty IoU union, we assign zero, following
the multi-task metric convention. Because any empty ground-truth mask
also has zero foreground IoU when predictions are nonempty, such ties
do not establish equal false-positive behavior. We explicitly count
false positives on empty-target tiles.

\section{Quantitative Results}

\begin{table}[!ht]
  \caption{Complete-surface comparison on the common 903-image CHN6-CUG
  test partition. Best scores in each column are in bold.}
  \label{tab:complete}
  \centering
  \small
  \setlength{\tabcolsep}{4pt}
  \begin{tabular}{lrrrr}
    \toprule
    Model & IoU & F1 & Prec. & Rec. \\
    \midrule
    Binary SegFormer-B4 & 64.53 & 78.44 & 74.51 & \textbf{82.80} \\
    Multi-class (ours) & \textbf{65.53} & \textbf{79.18} & \textbf{76.65} & 81.88 \\
    \bottomrule
  \end{tabular}
\end{table}

\subsection{Complete Surface}

Table~\ref{tab:complete} reports the primary comparison. The multi-task
system obtains 65.53\% IoU and 79.18\% F1, compared with 64.53\% and
78.44\% for the binary baseline. Using the unrounded scores, the
absolute differences are +1.01 percentage points in IoU and +0.74
points in F1. Precision increases by 2.14 points, while recall decreases
by 0.93 points, giving a higher overall foreground-overlap score.

The corresponding confusion counts make the aggregate trade-off
concrete. Binary prediction produces 11,231,241 true positives,
3,841,249 false positives, and 2,332,686 false negatives. Multi-task
prediction produces 11,105,769 true positives, 3,382,519 false
positives, and 2,458,158 false negatives. It removes 458,730 false
positives, a reduction of 11.94\%, with 125,472 fewer true positives.
This net trade-off improves both IoU and F1. The visual comparisons
further show that the model can recover specific missed corridors
while reducing false foreground in other scenes.

\begin{table}[!ht]
  \caption{Class-wise results of the proposed model on the 903-image
  CHN6-CUG test partition. The output masks may overlap.}
  \label{tab:targets}
  \centering
  \small
  \setlength{\tabcolsep}{4pt}
  \begin{tabular}{lrrrr}
    \toprule
    Target & IoU & F1 & Prec. & Rec. \\
    \midrule
    Complete surface & 65.53 & 79.18 & 76.65 & 81.88 \\
    Centerline & 7.52 & 13.98 & 14.07 & 13.90 \\
    Regular surface & 61.06 & 75.83 & 75.56 & 76.10 \\
    Intersection surface & 30.54 & 46.79 & 49.40 & 44.44 \\
    \bottomrule
  \end{tabular}
\end{table}

\subsection{Target-Specific Behavior}

Table~\ref{tab:targets} reports each semantic output separately. The
class-wise visualization in Figure~\ref{fig:multiclass} shows how the
regular and intersection outputs describe different parts of the road
scene, which the binary baseline does not distinguish.

Centerline exact-pixel F1 is 13.98\%. With the specified two-pixel
tolerance, precision is 66.70\%, recall is 34.32\%, and F1 is 45.32\%.
The difference between exact and tolerant scores reflects the
importance of small spatial displacements for thin centerlines.
Width MAE is 2.5681 pixels over valid annotations. Together, these
measurements characterize geometric outputs beyond the complete
road mask.

\subsection{Distribution of Paired Outcomes}

Multi-task IoU exceeds the binary result on 292 images and is lower on
191. The remaining 420 ties include all 404 empty-target images under
the foreground-IoU convention. Across the 499 nonempty images, mean
per-image IoU changes from 58.01\% to 59.79\%, and the median paired
difference is +0.88 points. Across all images, mean paired difference
is +0.98 points and median difference is zero. These macro summaries
complement, rather than replace, the micro scores of
Table~\ref{tab:complete}.

On the 404 empty-target images, binary prediction produces 59,893
false-positive pixels on 10 images; multi-task prediction produces
55,596 on 9 images. These pixels are included in global IoU and
precision. Reporting them separately avoids treating the many
zero-IoU ties as proof of equal background suppression.

\section{Qualitative Results}

\begin{figure*}[!t]
  \centering
  \includegraphics[width=0.90\textwidth]{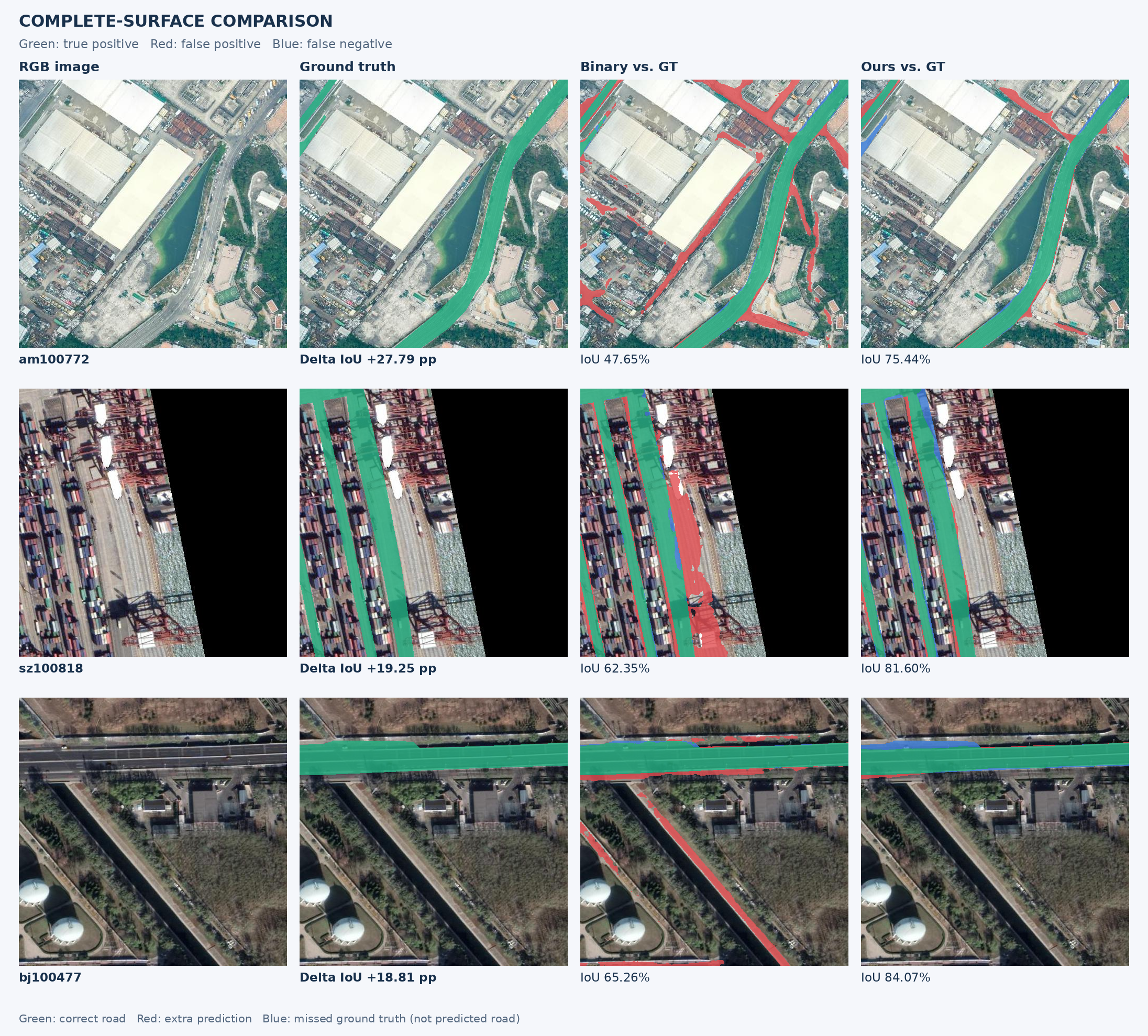}
  \caption{Complete-surface comparison on three selected scenes.
  Columns show the image, ground truth, binary agreement, and
  proposed-model agreement. Green: correctly predicted road;
  red: extra prediction; blue: missed ground truth, not predicted
  road. Both models have substantial overlap in every displayed case.}
  \Description{Three complete-surface comparison rows with image,
  ground truth, binary-baseline errors, and multi-task errors.}
  \label{fig:improvements}
\end{figure*}

\subsection{Disclosed Case Selection}

Figure~\ref{fig:improvements} compares three scenes with substantial
annotated road support and nonzero baseline overlap. We rank positive
paired IoU differences after requiring at least 15,000 road pixels,
baseline IoU at least 0.35, and proposed-model IoU at least 0.70,
then select three distinct region prefixes. Ties are broken by image
identifier. This retains meaningful binary detections while making
the additional refinement visible. These are selected qualitative
cases; the tables use all 903 images.

The comparison overlays use the same RGB image and target for both
systems. Green denotes true positives, red false positives, and blue
false negatives. No prediction is manually edited, locally
rethresholded, or morphologically repaired; displayed IoUs are
computed before rendering. The accompanying scripts record the
selection and rendering procedures.

\begin{figure*}[!t]
  \centering
  \includegraphics[width=0.92\textwidth]{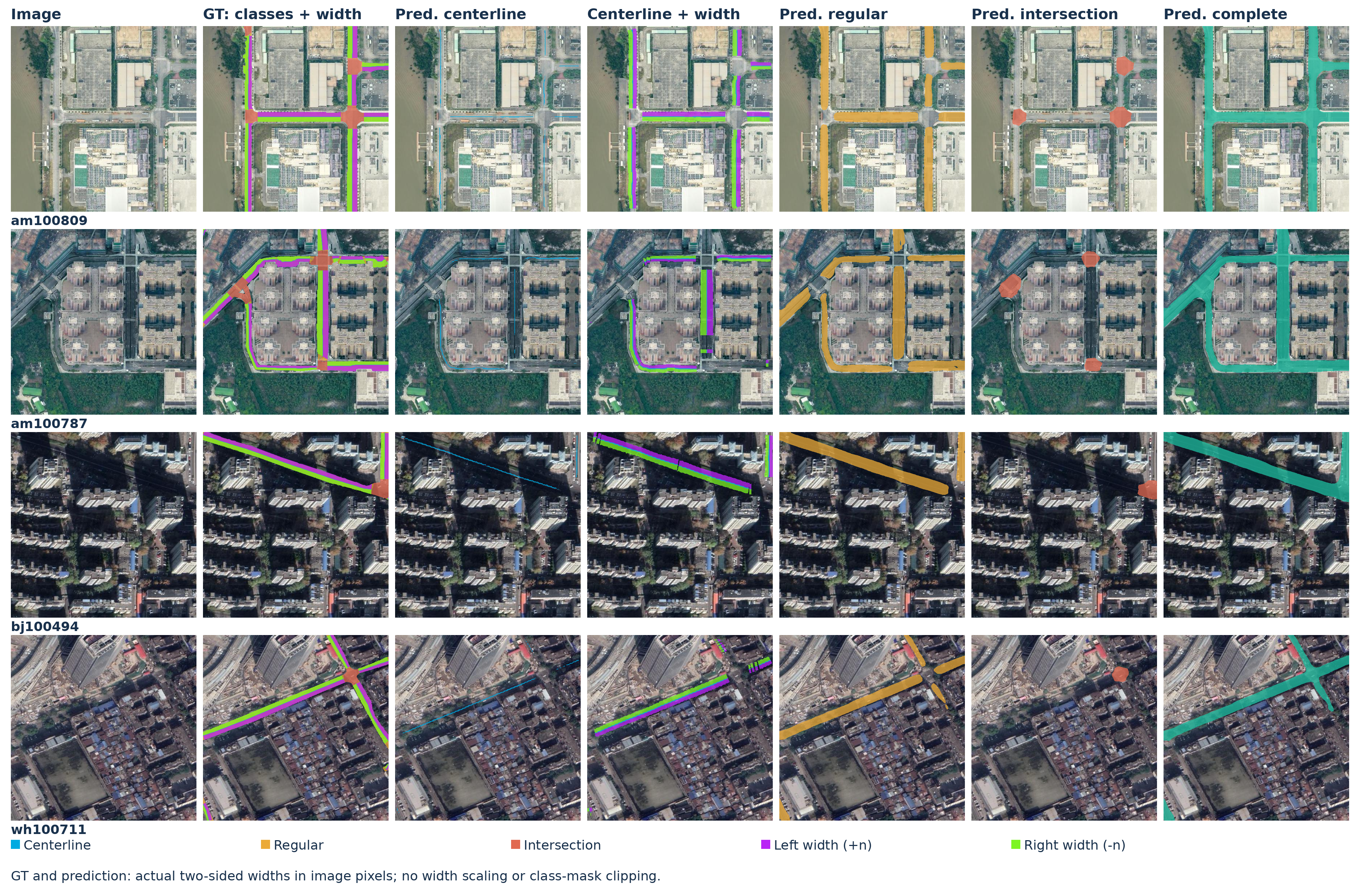}
  \caption{Multi-class predictions on four distinct test scenes.
  Columns show RGB, composite ground truth, predicted centerline,
  predicted centerline with width, regular surface, intersection,
  and complete surface. Ground truth combines regular (gold),
  intersection (coral), centerline (cyan), and valid width samples.
  Left/right width components are shown in purple/bright green at
  their actual pixel lengths, allowing overlap with semantic classes.
  Ground truth uses stored normals; prediction uses normals estimated
  from its own centerline without ground-truth conditioning.
  Cases are selected for joint
  output quality, not used to compute aggregate scores.}
  \Description{Four distinct scenes and seven columns showing the
  image, combined ground truth, and five prediction overlays including
  two-sided actual-width bands in purple and bright green.}
  \label{fig:multiclass}
\end{figure*}

\subsection{Suppressing False Road Extent}

In \texttt{am100772}, a road passes construction
and industrial-looking surroundings. The baseline recognizes the
main corridor but also assigns substantial nearby area to road.
Multi-task precision rises from 48.72\% to 81.38\%, with recall
changing from 95.63\% to 91.18\%. IoU consequently increases from
47.65\% to 75.44\%. The benefit here is predominantly suppression
of false foreground rather than recovery from an empty prediction.

In \texttt{sz100818}, the scene contains container stacks and broad
paved areas beside a comparatively narrow annotated road corridor.
The binary model extends foreground into the neighboring pavement.
The proposed output follows the annotated strip more closely,
reducing the wide band of false positives. IoU rises from 62.35\%
to 81.60\%, a gain of 19.25 points. This example separates road
surface extraction from indiscriminate paved-area recognition.

Tile \texttt{bj100477} contains a broad horizontal road and a diagonal
linear feature below it. Relative to the supplied complete-surface
target, the binary output has excess foreground along the diagonal
feature and at the main corridor boundary. The proposed output
suppresses much of that response while retaining the annotated road.
IoU improves from 65.26\% to 84.07\%. The interpretation is agreement
with the benchmark mask; it does not assert that every unannotated
paved feature is physically non-road.

Across these scenes, the visible advantage is better control of
surface extent where the binary baseline already detects roads.
The examples therefore complement the aggregate 11.94\% reduction
in false-positive pixels without relying on zero-overlap baseline
cases.

\subsection{Multi-Class Road Representations}

Figure~\ref{fig:multiclass} shows four different scenes rather than
repeating one scene across task rows. To select examples with useful
output in every class, we require at least 15,000 complete-surface
and 700 intersection pixels. We rank the mean and minimum of
complete IoU, tolerant centerline F1, regular IoU, and intersection
IoU with equal weight, then retain four distinct region prefixes.
The second case is curated for centerline continuity and visible
junctions, allowing a repeated region; selection records are provided.

The second column combines ground-truth surface classes and centerline
with width annotations in one reference panel. The separate predicted
centerline-plus-width column expands the two saved width components
along local normals at their actual lengths in image pixels.
Purple denotes the local left-side component, component 0 along the
positive path normal; bright green denotes the local right-side
component, component 1 along the negative normal. These sides are
relative to the oriented path, not compass directions. Ground truth
uses stored normals at valid annotations. Prediction estimates normals
from its skeletonized
centerline using the target generator's path-based rule with radius
8. Adjacent aligned cross sections are joined into bands without
clipping to semantic masks or scaling width values. No ground-truth
geometry is used to render prediction columns.

In \texttt{am100809}, elongated regular-road segments are separated
by localized intersection predictions, and the complete output
combines their spatial support. The centerline follows the road
axes without filling the whole paved width. In \texttt{am100787},
long centerline segments follow building blocks, with distinct
junction predictions.
In \texttt{bj100494}, the outputs describe a major road crossing
a shadowed residential scene, with junction support near the right
image edge. In \texttt{wh100711}, the regular branch follows the
dominant oblique corridor while the intersection branch localizes
the junction near its right end.

These overlays make the outputs interpretable in a common geographic
frame: longitudinal structure, non-junction surface, junction area,
and complete extent can be inspected separately on the same input.
They illustrate the intended joint representation; the class-wise
table remains the measure of performance over the full test set.

\section{Discussion and Limitations}

\subsection{Benefits for Road Surface Mapping}

The proposed model improves complete-surface IoU, F1, and precision
while providing regular-surface, intersection, centerline, and width
outputs. This extends binary extraction in two ways: better pooled
foreground overlap and a richer road representation. The visual
cases distinguish recovery of missing corridors from suppression
of spurious surrounding foreground; both contribute to practical
inspection of road maps.

\subsection{Scope and Future Work}

The comparison evaluates complete systems with different heads,
losses, and initial checkpoints; it does not isolate the effect of
one branch. The checkpoint-selection exposure is stated in the
experimental protocol. Matched ablations, repeated seeds, and
separate final model evaluation would strengthen the comparison.
The auxiliary-target construction and component ordering also need
to accompany a reproducible training release.

The present evaluation pools tiles from the six CHN6-CUG urban
areas. Spatially disjoint regional experiments and graph-level
evaluation are useful next steps for geographic transfer and
network applications. The current evidence concerns raster
segmentation and sparse width estimation.

\section{Conclusion}

We presented centerline-guided multi-class road surface segmentation,
with centerline conditioning, width-guided regular-surface decoding,
and learned surface refinement. On the common
903-image CHN6-CUG test partition, its complete-surface output
achieves 65.53\% IoU and 79.18\% F1, compared with 64.53\% and 78.44\%
for the binary baseline. The gains are accompanied by improved
precision and higher mean IoU on nonempty images. Selected visual
comparisons show refinement of annotated road extent and suppression
of false foreground. Class-wise results and multi-class
visualizations demonstrate the complementary road representations
produced by the same model.

\bibliographystyle{ACM-Reference-Format}
\bibliography{references}

\end{document}